\pdfoutput=1

\documentclass[11pt]{article}

\usepackage{EMNLP2023}

\usepackage{times}
\usepackage{latexsym}
\usepackage{graphicx}
\usepackage{tcolorbox}

\usepackage{tabularx}

\usepackage[T1]{fontenc}

\usepackage[utf8]{inputenc}

\usepackage{microtype}

\usepackage{inconsolata}

\title{Benchmarking the Medical Understanding and Reasoning of Large Language Models in Arabic Healthcare Tasks}

\author{Nouar AlDahoul \\
  Computer Science Department \\
   New York University \\
  Abu Dhabi, UAE \\
  \texttt{nouar.aldahoul@nyu.edu} \\\And
  Yasir Zaki \\
  Computer Science Department \\
   New York University \\
  Abu Dhabi, UAE \\
  \texttt{yasir.zaki@nyu.edu} \\}

\begin{document}
\maketitle
\begin{abstract}
Recent progress in large language models (LLMs) has showcased impressive proficiency in numerous Arabic natural language processing (NLP) applications. Nevertheless, their effectiveness in Arabic medical NLP domains has received limited investigation. This research examines the degree to which state-of-the-art LLMs demonstrate and articulate healthcare knowledge in Arabic, assessing their capabilities across a varied array of Arabic medical tasks. We benchmark several LLMs using a medical dataset proposed in the Arabic NLP AraHealthQA challenge in MedArabiQ2025 track. Various base LLMs were assessed on their ability to accurately provide correct answers from existing choices in multiple-choice questions (MCQs) and fill-in-the-blank scenarios. Additionally, we evaluated the capacity of LLMs in answering open-ended questions aligned with expert answers.
Our results reveal significant variations in correct answer prediction accuracy and low variations in semantic alignment of generated answers, highlighting both the potential and limitations of current LLMs in Arabic clinical contexts. 
Our analysis shows that for MCQs task, the proposed majority voting solution, leveraging three base models (Gemini Flash 2.5, Gemini Pro 2.5, and GPT o3), outperforms others, achieving up to 77\% accuracy and securing first place overall in the challenge~\cite{Leaderboard_AraHealth2025}. Moreover, for the open-ended questions task, several LLMs were able to demonstrate excellent performance in terms of semantic alignment and achieve a maximum BERTScore of 86.44\%.

\end{abstract}

\section{Introduction}

Medicine relies heavily on complex reasoning, spanning tasks from diagnostic decision-making to treatment planning, especially when patient outcomes depend on understanding multi-factorial conditions~\cite{qiu2024towards,huang2025o1}. Differential diagnosis involves generating and narrowing down possible diagnoses using clinical evidence, requiring both extensive medical knowledge and logical reasoning to evaluate multiple hypotheses.

LLMs have demonstrated superior performance across various domains and applications, such as article debiasing~\cite{kuo2025neutralizing}, content moderation~\cite{aldahoul2024advancing}, and political leaning detection~\cite{aldahoul2024polytc}. In the healthcare domain, LLMs are reshaping the landscape of healthcare by transforming the way consultations, diagnoses, and treatment plans are delivered~\cite{yang2023large}. They offer new avenues for improving patient education through dynamic, conversational interactions, thereby enhancing both accessibility and patient autonomy. Beyond direct patient care, LLMs also show promise in supporting medical training and streamlining administrative responsibilities, including the generation of clinical notes, referral letters, and discharge summaries~\cite{yang2023large}.

Most existing benchmarks focus on English, leaving a gap in evaluating Arabic LLMs for healthcare due to the lack of high-quality clinical datasets, Arabic’s linguistic diversity, and the limited performance of multilingual models in domain-specific tasks~\cite{daoud2025medarabiq}. To fill these gaps, there is an increasing demand for frameworks that evaluate LLM performance in clinical tasks for Arabic-speaking communities. Our analyses and experiments center around the
following research questions: \textbf{RQ1}: Do state-of-the-art proprietary base LLMs perform well in Arabic medical tasks? \textbf{RQ2}: To what extent do state-of-the-art proprietary base LLMs with reasoning capacity excel in Arabic medical tasks? \textbf{RQ3}: Do open-source-based Arabic LLMs perform well in Arabic medical tasks? and \textbf{RQ4}: How does majority voting among several LLMs enhance performance in Arabic medical tasks?

We address \textbf{RQ1} by running the APIs of several LLMs, such as Claude Opus, Grok 3, Deepseek v3, Llama 4 Maverick, GPT-4o-mini, and GPT-4o. To answer \textbf{RQ2}, we utilized APIs of state-of-the-art LLMs with reasoning capabilities such as GPT-o3, Gemini Flash 2.5, and Gemini Pro 2.5. Moreover, to address \textbf{RQ3}, we ran Falcon 3, Fanar, and Allam. Additionally, to answer \textbf{RQ4}, we calculated the majority vote among the predictions of three LLMs.

\section{Related Work}

BioBERT~\cite{lee2020biobert}, SCIBERT~\cite{beltagy2019scibert}, and PubMedBERT~\cite{gu2021domain} improved biomedical NLP by training on domain-specific corpora, thereby outperforming the general BERT model~\cite{yang2023large}. Building on this, ClinicalBERT~\cite{alsentzer2019publicly} enhanced performance on medical tasks by fine-tuning BERT and BioBERT using the MIMIC-III clinical dataset. Expanding further, GatorTrona~\cite{yang2022large}. significantly larger model trained from scratch on extensive clinical and biomedical text—demonstrated strong results across a wide range of clinical NLP tasks~\cite{yang2023large}.

Various benchmarks have been developed to evaluate LLMs' proficiency in medical reasoning and knowledge~\cite{huang2025o1,zuo2025medxpertqa}. However, significant challenges persist, ranging from ethical and safety concerns to the risk of biased outputs and inconsistent performance across different languages and cultural settings~\cite{yang2023large,nazi2024large,daoud2025medarabiq}.

To advance medical LLMs, researchers have increasingly focused on creating multilingual medical datasets~\cite{qiu2024towards}. They introduced MMedC, a 25.5-billion-token multilingual medical corpus, and MMedBench, a multilingual QA benchmark with rationales. By fine-tuning Llama 3 (8B), they found it outperformed all other open-source models and approached GPT-4 performance. However, Arabic was not one of the languages included~\cite{qiu2024towards}.

Arabic medical benchmarks are limited and mostly focused on question-answering tasks. While resources like MMLU~\cite{hendrycks2020measuring}, AraSTEM~\cite{mustapha2024arastem}, and AraMed~\cite{alasmari2024aramed} offer valuable contributions, they do not fully cover the breadth of Arabic medical tasks, highlighting the need for more comprehensive benchmarking efforts. The previous issue was addressed by the MedArabiQ benchmark~\cite{daoud2025medarabiq}.

\section{Materials and Methods}

\subsection{Dataset Overview}

\begin{figure}
    \centering
    \includegraphics[width=1\linewidth]{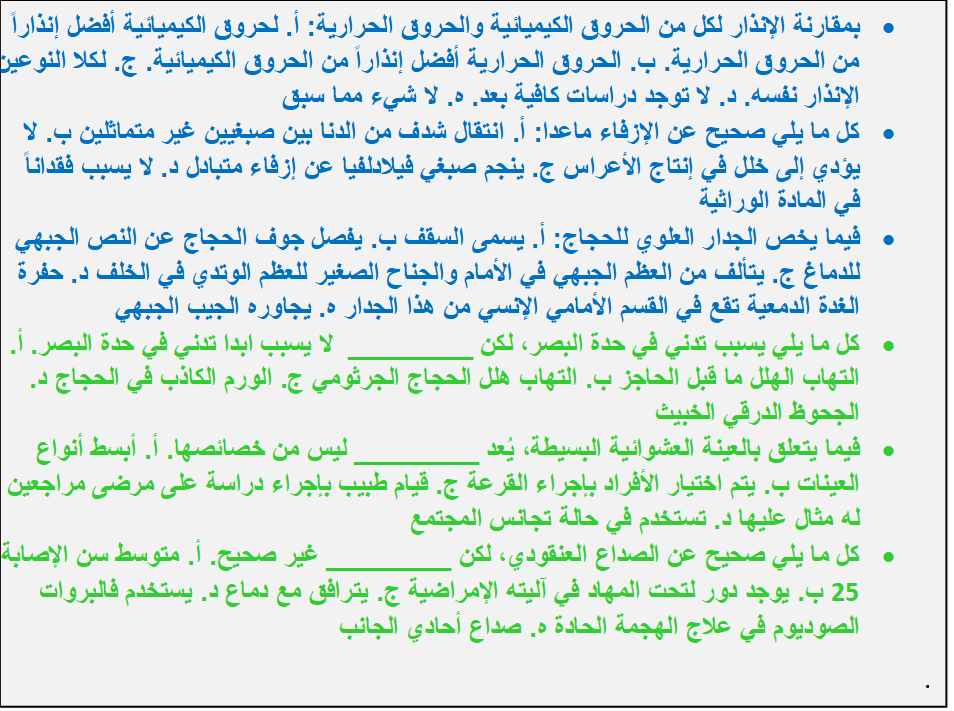}
    \caption{examples of multiple choice question answering (classification) dataset: MCQs in blue, Fill-in-the-blank questions with choices in green}
    \label{fig:dataset1}
\end{figure}

The medical data used in this work is the main dataset utilized in the AraHealthQA shared task in the MedArabiQ2025 track under one of the Arabic NLP challenges. It focuses on modern standard Arabic (MSA) and consists of 700 diverse clinical samples, covering both structured medical knowledge assessments and real-world patient-doctor interactions~\cite{daoud2025medarabiq}. The first dataset is for multiple choice question answering (classification). Few examples of the first dataset are shown in Figure~\ref{fig:dataset1}. The dataset is distributed as follows:

\begin{itemize}
\item a random set of 100 multiple-choice questions to evaluate the models’ medical understanding.
\item a set of 100 multiple-choice questions with bias injected to evaluate how LLMs handle ethical or culturally sensitive scenarios.
\item  a set of 100 fill-in-the-blank questions with choices to evaluate the model’s ability to recognize correct answers, reducing
the reliance on generative capabilities.

\end{itemize}

\begin{figure}
    \centering
    \includegraphics[width=1\linewidth]{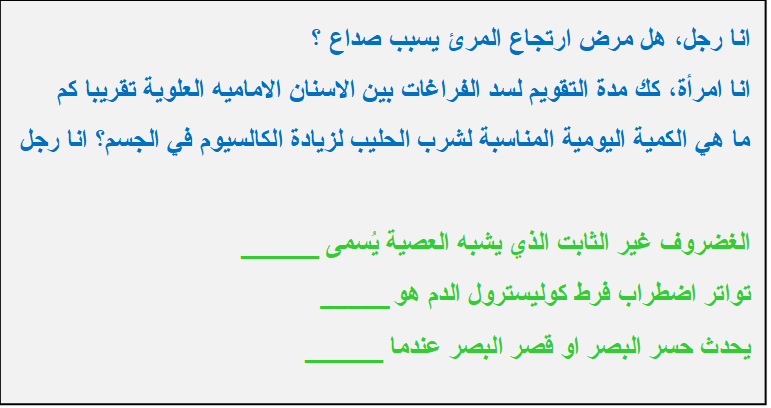}
    \caption{examples of open-ended question dataset: patient-doctor Q\&A in blue, Fill-in-the-blank without choices in green}
    \label{fig:dataset2}
\end{figure}

The second dataset is 
for open-ended question answering (generation). Few examples of the second dataset are shown in Figure~\ref{fig:dataset2}. The dataset is distributed as follows:

\begin{itemize}
\item a set of 100 fill-in-the-blank questions without choices to assess LLMs' reasoning and generation capabilities.
\item a set of 100 patient-doctor Q\&As selected from AraMed~\cite{alasmari2024aramed} to evaluate LLMs with online real-world scenarios from medical discussion forums.
\item a 100 Q\&As with grammatical error
correction to handle inflectional patterns and prepare the dataset for grammatical correction.
\item  a 100 Q\&As with LLM Modifications
 to mitigate potential model memorization and to assess the model's reasoning and adaptability.
\end{itemize}

The previous 700 examples were used for evaluation of LLMs. Later, another set of 200 examples (100 MCQs and 100 open-ended questions) was released for testing the LLMs' reasoning and understanding.

\subsection{Methods}

We have evaluated state-of-the-art base LLMs to identify the best in terms of correct answer match accuracy in MCQs task and alignment score of generated answers in open-ended questions task. This LLM can understand the questions, identify the correct answers utilizing its embedded knowledge and reasoning capability, and generate the answers that align with those of experts. 

We started assessing several proprietary base LLMs for the MCQs task to evaluate the accuracy of the match between real and predicted answers. We used LLMs' APIs in the inference mode utilizing two different zero-shot prompts specialized for the MCQs task (Prompt 1 and Prompt 2). The evaluated LLMs are:  Gemini Flash 2.5~\cite{team2023gemini,gemini_25_thinking}, Gemini Pro 2.5~\cite{team2023gemini,gemini_25_thinking}, GPT-4o-mini~\cite{GPT-4o}, GPT-4o~\cite{GPT-4o}, GPT o3~\cite{openai2025o3o4mini}, Grok 3~\cite{grok3}, Claude Opus~\cite{claude}, Deepseek v3~\cite{deepseek2024}, and Llama 4, and Maverick~\cite{Llama4}.

\begin{figure}
    \centering
    \includegraphics[width=1\linewidth]{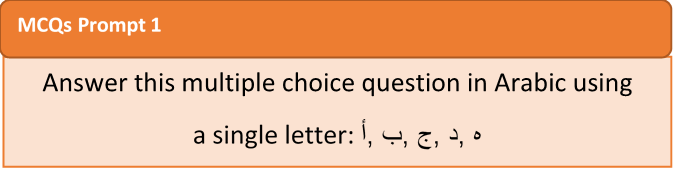}
  \label{fig:MCQs_Prompt1}
\end{figure}

\begin{figure}
    \centering
    \includegraphics[width=1\linewidth]{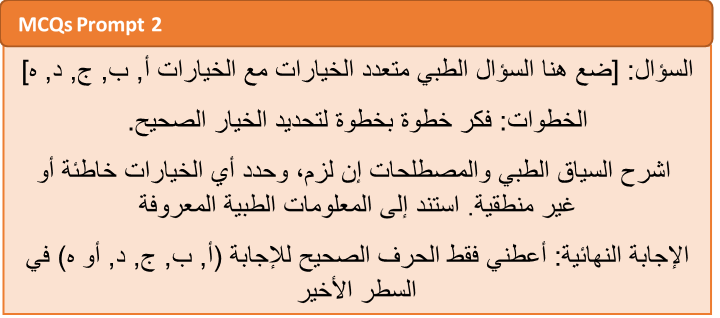}
 \label{fig:MCQs_Prompt2}
\end{figure}

Later, we selected the two LLMs that have shown high performance in the MCQs task: Gemini Flash 2.5 and Gemini Pro 2.5 and utilized them in the open-ended question task. We also demonstrated the performance of small-sized LLMs such as GPT-4o-mini in this task. We utilized three different prompts specialized for open-ended tasks (Prompt 1, Prompt 2, and Prompt 3).

\begin{tcolorbox}[colback=red!5!white,colframe=red!75!black, title={Open-ended questions' Prompt 1}, rounded corners, boxrule=1pt, boxsep=1pt]
You are a knowledgeable and concise medical expert. Provide a high-quality answer to the following open-ended medical question.\\
Your response should:\\
Begin with a direct, evidence-based answer.\\
Elaborate on the mechanisms, relevant anatomy or physiology, and clinical significance.\\
Use clear, professional medical language.\\
Question:\\
{[Insert your medical question here]}

\end{tcolorbox}

\begin{tcolorbox}[colback=blue!5!white,colframe=blue!75!black, title={Open-ended questions' Prompt 2}, rounded corners, boxrule=1pt, boxsep=1pt]
You are a knowledgeable and concise medical expert. Provide a high-quality answer to the following open-ended medical question.
\end{tcolorbox}

\begin{tcolorbox}[colback=green!5!white,colframe=green!75!black, title={Open-ended questions' Prompt 3}, rounded corners, boxrule=1pt, boxsep=1pt]
You are a knowledgeable and concise medical expert.\\
Your task is to generate a concise, accurate, and medically correct answer in Modern Standard Arabic.\\
Do not include explanations—just provide the best possible answer based on your knowledge.
\end{tcolorbox}

Additionally, open-source-based Arabic LLMs such as Falcon3~\cite{Falcon3,tiiuae_falcon3-7b-instruct} (``tiiuae/Falcon3-7B-Instruct''), Fanar~\cite{fanar1-9b-instruct,fanarllm2025} (``QCRI/Fanar-1-9B-Instruct''), and Allam~\cite{allam7b,bari2024allam} (``ALLaM-AI/ALLaM-7B-Instruct-preview'') were assessed for both tasks. 

We applied zero-shot prompting across all models and tasks, setting the temperature to 0 and top\_p to 1 for all tasks to ensure deterministic responses. For the open-ended question task, BERTScore was used as an evaluation metric to measure alignment between generated and expert answers. For this purpose, we used the "XLM-RoBERTa-Large model"~\cite{daoud2025medarabiq}, which was trained on multiple languages, including Arabic.

We also evaluated Arabic Falcon~\cite{falcon-arabic}. Since there is no API available for Arabic Falcon, we used the web interface to manually input questions into the chat version. We retained the history of previous questions to avoid clearing the context before each new query.

\subsection{Results and Discussion}

The results of the MCQs task using the proprietary LLMs are shown in Table~\ref{tab:probrietary_llms}. The dataset has MCQs related to understanding and reasoning. While understanding involves factual knowledge, reasoning mimics how doctors make decisions.

The medical reasoning capacity of GPT-o3, Gemini Flash 2.5, and Gemini Pro 2.5 makes them have superior performance compared to other LLMs. These simulate diagnostic thinking by combining multiple facts and using step-by-step reasoning to eliminate plausible but incorrect distractors in medical MCQs, which answers \textbf{RQ2}.

\begin{table}[h]
\centering
\begin{tabularx}{\columnwidth}{m{3cm}XX}
\hline
\textbf{Model}                       & \textbf{Prompt} & \textbf{Accuracy\%} \\ \hline
\textbf{GPT-4o-mini}                 & 1               & 49                \\ 
\textbf{GPT-4o}                      & 1               & 57                 \\ 
\textbf{GPT-O3}                      & 1               & 72                \\ 
\textbf{Gemini Flash   2.5}          & 1               & 73               \\ 
\textbf{Gemini Pro   2.5}            & 1               & 75                \\ 
\textbf{GPT-O3}                      & 2               & 74                \\
\textbf{Gemini Flash   2.5}          & 2               & 74                 \\ 
\textbf{Gemini Pro   2.5}            & 2               & 76                 \\ 
\textbf{Majority voting} & 2               & 77                \\ 
\textbf{Grok 3}                      & 2               & 60                 \\ 
\textbf{Claude Opus}                 & 2               & 49                 \\
\textbf{Falcon Arabic}               & 2               & 38                 \\ 
\textbf{Deepseek v3}                 & 2               & 56                 \\ 
\textbf{Llama 4  Maverick}            & 2               & 63                \\ \hline
\end{tabularx}
\caption{Accuracy of different proprietary base LLMs using different prompts.}
\label{tab:probrietary_llms}
\end{table}

Even though Claude 3, Deepseek 3, Grok 3, and Llama 4 Maverick possess strong reasoning capabilities, they exhibit modest performance on this task, likely due to limited medical knowledge or insufficient proficiency in Arabic, which addresses \textbf{RQ1} and \textbf{RQ2}. However, Llama 4 Maverick was the best among them in terms of accuracy (63\%).

For sensitivity of prompt construction, we found that Prompt 2, which includes step-by-step or chain-of-thought reasoning, is generally better than simple Prompt 1 when it comes to answering medical MCQs. 

The significant finding in this work is that current state-of-the-art proprietary LLMs exhibit limitations in their embedded medical knowledge of various Arabic medical tasks (maximum accuracy is 76\% in Gemini Pro 2.5). The source of errors in the MCQ task may stem from misunderstanding of questions, lack of medical knowledge, or lack of medical reasoning capabilities.

To benefit from the capacity of each of three LLMs (GPT-O3, Gemini Flash 2.5, and Gemini Pro 2.5) in MCQs task, we applied a majority voting technique using the predictions from these LLMs, resulting in a final accuracy of 77\%, which secured first place overall in the challenge, which answers \textbf{RQ4}.

The results of the open-ended questions task using proprietary LLMs are shown in Table~\ref{tab:open-ended}. The dataset has questions labeled with answers. The LLMs should generate answers that are semantically aligned with reference answers.

Our finding indicates that reasoning LLMs such as Gemini Flash 2.5 and Gemini Pro 2.5 have structured answers that reduce hallucination and overconfidence, as the models are less likely to guess and more likely to justify their answers. As a result, their responses often align more closely with reference answers and perform better on semantic evaluation metrics like BERTScore, which answers \textbf{RQ2}. Furthermore, GPT-4o-mini shows good performance in terms of BERTScore.

Additionally, the three LLMs showed high sensitivity to prompts with variances in BERTScores. The maximum BERTScores were achieved by Prompt 3 that asked the LLMs to have modern standard Arabic in response, emphasized medically correct answers, and asked for concise answers that are not diluted with explanations, which usually tend to align more closely with reference answers.

\begin{table}[h]
\centering
\begin{tabular}{ccc}
\hline
\textbf{Model}              & \textbf{Prompt} & \textbf{BERTScore} \\ \hline
\textbf{Gemini Pro   2.5}   & 1               & 0.8105                 \\ 
\textbf{Gemini Flash   2.5} & 2               & 0.8364                 \\ 
\textbf{GPT-4o-mini}        & 2               & 0.8386                 \\ 
\textbf{GPT-4o-mini}        & 3               & 0.8581   \\
\textbf{Gemini Flash   2.5} & 3               & 0.8633           
              \\
\textbf{Gemini Pro   2.5}   & 3               & 0.8644                 \\ 
 \hline
\end{tabular}
\caption{\normalsize BERTScore of proprietary  base LLMs using different prompts.}
\label{tab:open-ended}
\end{table}

Table~\ref{tab:arabic} shows the accuracy and BERTScore of several open-source base Arabic LLMs. Among the models, Allam demonstrates relatively better performance (39\%) in MCQs task, while Falcon 3 gave the best BERTScore (0.8493). This experiment indicates a lack of medical knowledge and/or medical reasoning in the base open-source Arabic LLMs compared to proprietary ones, which
addresses \textbf{RQ3}.

\begin{table}[h]
\centering
\begin{tabular}{ccc}
\hline
\textbf{Model}                       & \textbf{Task} & \textbf{Accuracy \% } \\ \hline
\textbf{Falcon 3}                           & Task 1     & 36                \\ 
\textbf{Fanar}                            & Task 1        & 31                 \\ 
\textbf{Allam}                             & Task 1         & 39                \\   \hline
\textbf{Model}                       & \textbf{Task} & \textbf{BERTScore} \\ \hline

\textbf{Falcon 3}                          & Task 2       & 0.8493                \\ 
\textbf{Fanar}                             & Task 2      & 0.8403                 \\ 
\textbf{Allam}                              & Task 2      & 0.8431               \\ 
\hline
\end{tabular}
\caption{Accuracy and BERTScore of different base Arabic LLMs.}
\label{tab:arabic}
\end{table}

\section*{Limitations}

The first limitation is that multiple-choice and fill-in-the-blank with choice questions in the MedArabiQ2025 dataset are limited to only a few hundred examples. There is a clear need for larger, high-quality Arabic medical datasets to fine-tune LLMs and enhance their performance. Alternatively, storing extensive medical data in a vector database and employing retrieval-augmented generation (RAG) techniques could help retrieve more accurate and contextually relevant answers.

A second limitation of this work is the absence of bias detection and mitigation techniques during the preprocessing of questions before inputting them to LLMs. Incorporating such techniques could play a significant role in improving model performance and ensuring more reliable outputs.

The third limitation is that for open-ended and fill-in-the-blank questions without choices, we lack a robust metric for capturing semantic similarity. In this work, we utilized BERTScore, which often yields similar values across different responses and fails to reflect subtle nuances in semantic alignment with the correct answers.

\bibliography{anthology,custom}

\begin{thebibliography}{33}
\expandafter\ifx\csname natexlab\endcsname\relax\def\natexlab#1{#1}\fi

\bibitem[{Lea()}]{Leaderboard_AraHealth2025}

\newblock Arahealthqa 2025 shared task - track 2 (sub-task 1).
\newblock \url{https://www.codabench.org/competitions/8967/#/results-tab}.
\newblock Accessed: 2025‑08‑05.

\bibitem[{cla(2024)}]{claude}
 2024.
\newblock Using the api getting started.
\newblock \url{https://docs.anthropic.com/en/api/getting-started}.

\bibitem[{Alasmari et~al.(2024)Alasmari, Alhumoud, and Alshammari}]{alasmari2024aramed}
Ashwag Alasmari, Sarah Alhumoud, and Waad Alshammari. 2024.
\newblock Aramed: Arabic medical question answering using pretrained transformer language models.
\newblock In \emph{Proceedings of the 6th Workshop on Open-Source Arabic Corpora and Processing Tools (OSACT) with Shared Tasks on Arabic LLMs Hallucination and Dialect to MSA Machine Translation@ LREC-COLING 2024}, pages 50--56.

\bibitem[{AlDahoul et~al.(2024{\natexlab{a}})AlDahoul, Rahwan, and Zaki}]{aldahoul2024polytc}
Nouar AlDahoul, Talal Rahwan, and Yasir Zaki. 2024{\natexlab{a}}.
\newblock Polytc: a novel bert-based classifier to detect political leaning of youtube videos based on their titles.
\newblock \emph{Journal of Big Data}, 11(1):80.

\bibitem[{AlDahoul et~al.(2024{\natexlab{b}})AlDahoul, Tan, Kasireddy, and Zaki}]{aldahoul2024advancing}
Nouar AlDahoul, Myles Joshua~Toledo Tan, Harishwar~Reddy Kasireddy, and Yasir Zaki. 2024{\natexlab{b}}.
\newblock Advancing content moderation: Evaluating large language models for detecting sensitive content across text, images, and videos.
\newblock \emph{arXiv preprint arXiv:2411.17123}.

\bibitem[{{ALLaM-AI}(2025)}]{allam7b}
{ALLaM-AI}. 2025.
\newblock Allam-7b-instruct-preview.
\newblock \url{https://huggingface.co/ALLaM-AI/ALLaM-7B-Instruct-preview}.
\newblock Accessed: 2025-07-13.

\bibitem[{Alsentzer et~al.(2019)Alsentzer, Murphy, Boag, Weng, Jin, Naumann, and McDermott}]{alsentzer2019publicly}
Emily Alsentzer, John~R Murphy, Willie Boag, Wei-Hung Weng, Di~Jin, Tristan Naumann, and Matthew McDermott. 2019.
\newblock Publicly available clinical bert embeddings.
\newblock \emph{arXiv preprint arXiv:1904.03323}.

\bibitem[{Bari et~al.(2024)Bari, Alnumay, Alzahrani, Alotaibi, Alyahya, AlRashed, Mirza, Alsubaie, Alahmed, Alabduljabbar et~al.}]{bari2024allam}
M~Saiful Bari, Yazeed Alnumay, Norah~A Alzahrani, Nouf~M Alotaibi, Hisham~A Alyahya, Sultan AlRashed, Faisal~A Mirza, Shaykhah~Z Alsubaie, Hassan~A Alahmed, Ghadah Alabduljabbar, et~al. 2024.
\newblock Allam: Large language models for arabic and english.
\newblock \emph{arXiv preprint arXiv:2407.15390}.

\bibitem[{Beltagy et~al.(2019)Beltagy, Lo, and Cohan}]{beltagy2019scibert}
Iz~Beltagy, Kyle Lo, and Arman Cohan. 2019.
\newblock Scibert: A pretrained language model for scientific text.
\newblock \emph{arXiv preprint arXiv:1903.10676}.

\bibitem[{Daoud et~al.(2025)Daoud, Abouzahir, Kharouf, Al-Eisawi, Habash, and Shamout}]{daoud2025medarabiq}
Mouath~Abu Daoud, Chaimae Abouzahir, Leen Kharouf, Walid Al-Eisawi, Nizar Habash, and Farah~E Shamout. 2025.
\newblock Medarabiq: Benchmarking large language models on arabic medical tasks.
\newblock \emph{arXiv preprint arXiv:2505.03427}.

\bibitem[{DeepSeek(2024)}]{deepseek2024}
DeepSeek. 2024.
\newblock Deepseek chat.
\newblock \url{https://www.deepseek.com}.
\newblock Accessed: July 13, 2025.

\bibitem[{Gu et~al.(2021)Gu, Tinn, Cheng, Lucas, Usuyama, Liu, Naumann, Gao, and Poon}]{gu2021domain}
Yu~Gu, Robert Tinn, Hao Cheng, Michael Lucas, Naoto Usuyama, Xiaodong Liu, Tristan Naumann, Jianfeng Gao, and Hoifung Poon. 2021.
\newblock Domain-specific language model pretraining for biomedical natural language processing.
\newblock \emph{ACM Transactions on Computing for Healthcare (HEALTH)}, 3(1):1--23.

\bibitem[{Hendrycks et~al.(2020)Hendrycks, Burns, Basart, Zou, Mazeika, Song, and Steinhardt}]{hendrycks2020measuring}
Dan Hendrycks, Collin Burns, Steven Basart, Andy Zou, Mantas Mazeika, Dawn Song, and Jacob Steinhardt. 2020.
\newblock Measuring massive multitask language understanding.
\newblock \emph{arXiv preprint arXiv:2009.03300}.

\bibitem[{Huang et~al.(2025)Huang, Geng, Hua, Huang, Zou, Zhang, Liu, and Zhang}]{huang2025o1}
Zhongzhen Huang, Gui Geng, Shengyi Hua, Zhen Huang, Haoyang Zou, Shaoting Zhang, Pengfei Liu, and Xiaofan Zhang. 2025.
\newblock O1 replication journey--part 3: Inference-time scaling for medical reasoning.
\newblock \emph{arXiv preprint arXiv:2501.06458}.

\bibitem[{{Koray Kavukcuoglu}(2025)}]{gemini_25_thinking}
{Koray Kavukcuoglu}. 2025.
\newblock Gemini 2.5: Our most intelligent ai model.
\newblock \url{https://blog.google/technology/google-deepmind/gemini-model-thinking-updates-march-2025/#gemini-2-5-thinking}.

\bibitem[{Kuo et~al.(2025)Kuo, Chu, AlDahoul, Ibrahim, Rahwan, and Zaki}]{kuo2025neutralizing}
Chen~Wei Kuo, Kevin Chu, Nouar AlDahoul, Hazem Ibrahim, Talal Rahwan, and Yasir Zaki. 2025.
\newblock Neutralizing the narrative: Ai-powered debiasing of online news articles.
\newblock \emph{arXiv preprint arXiv:2504.03520}.

\bibitem[{Lee et~al.(2020)Lee, Yoon, Kim, Kim, Kim, So, and Kang}]{lee2020biobert}
Jinhyuk Lee, Wonjin Yoon, Sungdong Kim, Donghyeon Kim, Sunkyu Kim, Chan~Ho So, and Jaewoo Kang. 2020.
\newblock Biobert: a pre-trained biomedical language representation model for biomedical text mining.
\newblock \emph{Bioinformatics}, 36(4):1234--1240.

\bibitem[{Mustapha et~al.(2024)Mustapha, Al-Khansa, Al-Mubasher, Mourad, Hamoud, El-Husseini, Al-Sakkaf, and Awad}]{mustapha2024arastem}
Ahmad Mustapha, Hadi Al-Khansa, Hadi Al-Mubasher, Aya Mourad, Ranam Hamoud, Hasan El-Husseini, Marwah Al-Sakkaf, and Mariette Awad. 2024.
\newblock Arastem: A native arabic multiple choice question benchmark for evaluating llms knowledge in stem subjects.
\newblock \emph{arXiv preprint arXiv:2501.00559}.

\bibitem[{Nazi and Peng(2024)}]{nazi2024large}
Zabir~Al Nazi and Wei Peng. 2024.
\newblock Large language models in healthcare and medical domain: A review.
\newblock In \emph{Informatics}, volume~11, page~57. MDPI.

\bibitem[{OpenAI(2024)}]{GPT-4o}
OpenAI. 2024.
\newblock \href {https://platform.openai.com/docs/models/gpt-4o} {Gpt-4o}.

\bibitem[{{OpenAI}(2025)}]{openai2025o3o4mini}
{OpenAI}. 2025.
\newblock Introducing o3 and o4-mini.
\newblock \url{https://openai.com/index/introducing-o3-and-o4-mini/}.
\newblock Accessed: 2025-07-13.

\bibitem[{{QCRI}(2025)}]{fanar1-9b-instruct}
{QCRI}. 2025.
\newblock Fanar‑1‑9b‑instruct.
\newblock \url{https://huggingface.co/QCRI/Fanar-1-9B-Instruct}.
\newblock Accessed: 2025-07-13.

\bibitem[{Qiu et~al.(2024)Qiu, Wu, Zhang, Lin, Wang, Zhang, Wang, and Xie}]{qiu2024towards}
Pengcheng Qiu, Chaoyi Wu, Xiaoman Zhang, Weixiong Lin, Haicheng Wang, Ya~Zhang, Yanfeng Wang, and Weidi Xie. 2024.
\newblock Towards building multilingual language model for medicine.
\newblock \emph{Nature Communications}, 15(1):8384.

\bibitem[{Team(2025{\natexlab{a}})}]{falcon-arabic}
Falcon-LLM Team. 2025{\natexlab{a}}.
\newblock \href {https://falcon-lm.github.io/blog/falcon-arabic} {Falcon-arabic: A breakthrough in arabic language models}.

\bibitem[{Team et~al.()Team, Abbas, Ahmad, Alam, Altinisik, Asgari, Boshmaf, Boughorbel, Chawla, Chowdhury, Dalvi, Darwish, Durrani, Elfeky, Elmagarmid, Eltabakh, Fatehkia, Fragkopoulos, Hasanain, Hawasly, Husaini, Jung, Lucas, Magdy, Messaoud, Mohamed, Mohiuddin, Mousi, Mubarak, Musleh, Naeem, Ouzzani, Popovic, Sadeghi, Sencar, Shinoy, Sinan, Zhang, Ali, Kheir, Ma, and Ruan}]{fanarllm2025}
Fanar Team, Ummar Abbas, Mohammad~Shahmeer Ahmad, Firoj Alam, Enes Altinisik, Ehsannedin Asgari, Yazan Boshmaf, Sabri Boughorbel, Sanjay Chawla, Shammur Chowdhury, Fahim Dalvi, Kareem Darwish, Nadir Durrani, Mohamed Elfeky, Ahmed Elmagarmid, Mohamed Eltabakh, Masoomali Fatehkia, Anastasios Fragkopoulos, Maram Hasanain, Majd Hawasly, Mus'ab Husaini, Soon-Gyo Jung, Ji~Kim Lucas, Walid Magdy, Safa Messaoud, Abubakr Mohamed, Tasnim Mohiuddin, Basel Mousi, Hamdy Mubarak, Ahmad Musleh, Zan Naeem, Mourad Ouzzani, Dorde Popovic, Amin Sadeghi, Husrev~Taha Sencar, Mohammed Shinoy, Omar Sinan, Yifan Zhang, Ahmed Ali, Yassine~El Kheir, Xiaosong Ma, and Chaoyi Ruan.
\newblock Fanar: An arabic-centric multimodal generative ai platform.

\bibitem[{Team et~al.(2023)Team, Anil, Borgeaud, Wu, Alayrac, Yu, Soricut, Schalkwyk, Dai, Hauth et~al.}]{team2023gemini}
Gemini Team, Rohan Anil, Sebastian Borgeaud, Yonghui Wu, Jean-Baptiste Alayrac, Jiahui Yu, Radu Soricut, Johan Schalkwyk, Andrew~M Dai, Anja Hauth, et~al. 2023.
\newblock Gemini: a family of highly capable multimodal models.
\newblock \emph{arXiv preprint arXiv:2312.11805}.

\bibitem[{Team(2025{\natexlab{b}})}]{Llama4}
Meta Team. 2025{\natexlab{b}}.
\newblock \href {https://ai.meta.com/blog/llama-4-multimodal-intelligence/} {The llama 4 herd: The beginning of a new era of natively multimodal ai innovation}.

\bibitem[{Team(2024)}]{Falcon3}
TII Team. 2024.
\newblock The falcon 3 family of open models.

\bibitem[{{Technology Innovation Institute (TII) UAE}(2024)}]{tiiuae_falcon3-7b-instruct}
{Technology Innovation Institute (TII) UAE}. 2024.
\newblock Falcon3‑7b‑instruct.
\newblock \url{https://huggingface.co/tiiuae/Falcon3-7B-Instruct}.

\bibitem[{xAI(2025)}]{grok3}
xAI. 2025.
\newblock \href {https://x.ai/news/grok-3} {Grok 3 beta — the age of reasoning agents}.

\bibitem[{Yang et~al.(2023)Yang, Tan, Lu, Thirunavukarasu, Ting, and Liu}]{yang2023large}
Rui Yang, Ting~Fang Tan, Wei Lu, Arun~James Thirunavukarasu, Daniel Shu~Wei Ting, and Nan Liu. 2023.
\newblock Large language models in health care: Development, applications, and challenges.
\newblock \emph{Health Care Science}, 2(4):255--263.

\bibitem[{Yang et~al.(2022)Yang, Chen, PourNejatian, Shin, Smith, Parisien, Compas, Martin, Costa, Flores et~al.}]{yang2022large}
Xi~Yang, Aokun Chen, Nima PourNejatian, Hoo~Chang Shin, Kaleb~E Smith, Christopher Parisien, Colin Compas, Cheryl Martin, Anthony~B Costa, Mona~G Flores, et~al. 2022.
\newblock A large language model for electronic health records.
\newblock \emph{NPJ digital medicine}, 5(1):194.

\bibitem[{Zuo et~al.(2025)Zuo, Qu, Li, Chen, Zhu, Hua, Zhang, Ding, and Zhou}]{zuo2025medxpertqa}
Yuxin Zuo, Shang Qu, Yifei Li, Zhangren Chen, Xuekai Zhu, Ermo Hua, Kaiyan Zhang, Ning Ding, and Bowen Zhou. 2025.
\newblock Medxpertqa: Benchmarking expert-level medical reasoning and understanding.
\newblock \emph{arXiv preprint arXiv:2501.18362}.

\end{thebibliography}
\bibliographystyle{acl_natbib}

\end{document}